\documentclass{article}
\usepackage[utf8]{inputenc}

\usepackage{graphicx} 
\usepackage{hyperref} 
\usepackage{amsmath} 
\usepackage{amsfonts} 
\usepackage{amssymb} 
\usepackage{booktabs} 
\usepackage{multirow} 
\usepackage{array} 
\usepackage{caption} 
\usepackage{subcaption} 
\usepackage{float} 
\usepackage{color} 
\usepackage{xcolor} 
\usepackage{url} 
\usepackage{doi} 
\usepackage{cite} 
\usepackage{authblk} 
\usepackage{amsmath, amssymb, algorithm, algpseudocode}
\usepackage{float}
\newcommand{\myand}{\hspace{2em}}

\title{\textbf{Multivariate Time Series Anomaly Detection using DiffGAN Model}}
\author{Guangqiang Wu \thanks{College of Science, University of Shanghai for Science and Technology,
		Shanghai 200093, China (email: {\tt 1684626441@qq.com}).}
	\myand Fu~Zhang\thanks{Corresponding author. College of Science, University of Shanghai for Science and Technology,
		Shanghai 200093, China (email: {\tt fuzhang82@gmail.com}).
		F.~Zhang was partially supported 
		by the National Natural Science Foundation of China (No. 12071292).}}

\begin{document}
	
\maketitle

\begin{abstract}
	In recent years, some researchers have applied diffusion models to multivariate time series anomaly detection. The partial diffusion strategy, which depends on the diffusion steps, is commonly used for anomaly detection in these models. However, different diffusion steps have an impact on the reconstruction of the original data, thereby impacting the effectiveness of anomaly detection. To address this issue, we propose a novel method named DiffGAN, which adds a generative adversarial network component to the denoiser of diffusion model. This addition allows for the simultaneous generation of noisy data and prediction of diffusion steps. Compared to multiple state-of-the-art reconstruction models, experimental results demonstrate that DiffGAN achieves superior performance in anomaly detection.
\end{abstract}


\section{Introduction}
Anomaly detection in multivariate time series (MTS-AD), represents a significant research domain, dedicated to uncovering atypical behaviors across multiple dimensions of time-oriented datasets. MTS-AD plays an indispensable role in predictive maintenance across various industries by facilitating the real-time surveillance of complex data streams generated by an array of machinery sensors\cite{zonta2020predictive, nakamura2021anomaly, ren2020timeseries}. Through early identification of emerging problems, this approach not only mitigates the risk of unexpected breakdowns but also significantly boosts operational efficiency and enhances safety protocols. The importance of anomaly detection lies in its power to transform reactive maintenance practices into proactive strategies, ensuring smoother operations and reducing potential hazards.

Traditional methods, such as One-Class Support Vector Machines (OC-SVM)\cite{zhou2023conditional}, perform well in handling static data but are less adept at managing the temporal information inherent in time series data. In contrast, deep learning methods excel in MTS-AD due to their robust capabilities in modeling the temporal dimension. Reconstruction-based methods are a typical subset of these techniques, which learn the characteristics of normal data through an autoencoding framework and use the reconstruction error as an anomaly score during detection. Generative Adversarial Networks (GANs)\cite{li2019mad, zhou2019beatgan, geiger2020tadgan} represent a prominent reconstruction-based method, enhancing anomaly detection accuracy by adversarially generating time series data.

In recent years, Diffusion Models (DMs)\cite{ho2020denoising, song2020denoising} have made significant strides in image generation, not only producing high-quality images but also demonstrating superior mode coverage compared to GANs. Motivated by these advancements, researchers have begun exploring the application of DMs in MTS-AD, aiming to capture the temporal dynamics of data through diffusion processes, thereby improving AD performance. DMs achieve this by progressively adding noise to data through a series of forward diffusion steps, transforming the data distribution towards a Gaussian distribution. Subsequently, the reverse denoising process gradually removes the noise, restoring the original data characteristics. This mechanism aids the model in better understanding and learning the intrinsic structure of the data, especially in handling complex time series.

However, anomaly detection differs from generative tasks in its requirement to first add noise to the test data and then use the diffusion model for denoising. Researchers commonly adopt a partial diffusion strategy in this process, which depends on the diffusion steps. Therefore, how to quickly determine the optimal diffusion steps becomes a key issue.

In this paper, we propose a novel anomaly detection method that integrates Diffusion Model and Generative Adversarial Network, named DiffGAN. This method combines the denoiser with the generator and discriminator, using the generator to replace the forward noising process in diffusion model and the discriminator to predict the required steps for denoising. This approach overcomes the limitations of fixed diffusion steps in traditional diffusion models, enhancing the flexibility and adaptability of the model. Experimental results demonstrate that DiffGAN outperforms existing benchmarks in terms of detection accuracy on multiple MTS datasets, demonstrating its significant potential in the field of MTS-AD.

In summary, this paper makes the following contributions:
\begin{enumerate}
	\item We propose a novel anomaly detection method (named DiffGAN) for MTS-AD, which overcomes the limitations of fixed diffusion steps in traditional diffusion models.
	\item We re-examine the role of the discriminator, responsible for assessing the level of noise added to the data, which not only guides the generator to produce noisy data that facilitates reconstruction but also predicts the required steps for denoising. In other words, we regard the discriminator as a controller in the data encoding process.
	\item Compared to multiple state-of-the-art reconstruction models, our experimental results demonstrate the potential of DiffGAN for MTS-AD. The relevant code and datasets of this paper are publicly available at \href{https://github.com/guangqiangWu/DiffGAN}{DiffGAN}.
\end{enumerate}

\section{Related Work}

In this section, we briefly introduce the basic concepts of Generative Adversarial Network and Diffusion Model, and summarize previous works on multivariate time series anomaly detection.We categorize these methods into three main sections: classical methods, GAN-based methods, and Diffusion-based methods. The first part focuses on classical methods, and the latter two parts delve into reconstruction-based methods, specifically those utilizing GANs and DMs.

\subsection{Classical Methods}
\label{sec:classical}
Classical Methods can be systematically categorized into three primary classes: distance-based, clustering-based, and probability-based. Distance-based algorithms, such as K-Nearest Neighbor (KNN)\cite{guo2018knn}, operate on the principle that the distance between an anomaly and a normal point is greater than the distance between normal points. While these methods are straightforward and effective for low-dimensional data, they are sensitive to noise and the selection of parameters. Clustering-based algorithms, including K-Means\cite{yin2015improved}, Clustering-Based Local Outlier Factor (CBLOF)\cite{ali2017detecting}, and Density-Based Spatial Clustering of Applications with Noise (DBSCAN)\cite{jain2022modified}, identify anomalies as points that do not belong to dense clusters. These methods use density as a criterion to detect outliers, but they can be computationally expensive and slow to converge, especially with large datasets. Probability-based algorithms, such as Deep Autoencoding Gaussian Mixture Model (DAGMM)\cite{zong2018deep}, fit the data to probabilistic models and flag points with low likelihood under the model as anomalies. GMMs are particularly useful for handling complex and multi-modal distributions, while DAGMM combines a deep autoencoder with a GMM to achieve superior performance in anomaly detection by leveraging a low-dimensional latent space.

\subsection{GAN-based Methods}

\subsubsection{Generative Adversarial Networks}
Generative Adversarial Networks (GANs), introduced by Goodfellow et al. in 2014 \cite{goodfellow2014generative}, are a class of machine learning models designed to generate new data samples that resemble the training data. A GAN consists of two main components: a generator ($ G $) and a discriminator ($ D $). The generator takes a random noise vector $ z $ as input and outputs synthetic data samples that aim to mimic the real data distribution. The discriminator, on the other hand, evaluates whether a given sample is real or fake, striving to accurately distinguish between genuine and synthesized data.

The training process of a GAN involves an adversarial game between the generator and the discriminator. The generator tries to produce samples that are indistinguishable from real data, while the discriminator aims to correctly classify the samples as either real or fake. This competition drives both components to improve iteratively. Mathematically, the objective function of a GAN can be formulated as:

\begin{equation}
	\min_G \max_D \mathbb{E}_{x \sim P_{\text{data}}}[ \log D(x) ] + \mathbb{E}_{z \sim P_z}[ \log (1 - D(G(z))) ],
\end{equation}
Here, $ x $ represents the real data samples drawn from the true data distribution $ P_{\text{data}} $, and $ z $ is a random noise vector sampled from a prior distribution $ P_z $. The discriminator $ D $ outputs a probability score indicating the likelihood that a given sample is real. The generator $ G $ aims to maximize the probability that the discriminator incorrectly classifies the generated samples as real. Through this adversarial training, the generator learns to produce highly realistic data, making GANs a powerful tool for various applications, including image generation, data augmentation, and anomaly detection.

\subsubsection{Anomaly Detection}

Recent advancements in GANs have led to significant interest in their application for MTS-AD. Despite the inherent temporal dependencies and complex patterns in MTS data, which make anomaly detection more challenging compared to image processing, several innovative approaches have emerged to address these issues. For instance, MAD-GAN\cite{li2019mad} proposes an unsupervised method that integrates the reconstruction error from the generator and the discriminative loss from the discriminator to compute an anomaly score. This dual-loss approach enhances the model's ability to identify subtle anomalies. Another notable method, BeatGAN\cite{zhou2019beatgan}, focuses on medical applications by comparing input heartbeats with generated heartbeats to pinpoint abnormal time points, aiding in the timely diagnosis and treatment of patients. TadGAN\cite{geiger2020tadgan}, another innovative method, introduces a cycle consistency loss into the GAN framework to improve the reconstruction of time series data. TadGAN utilizes both the generator and the critic to detect anomalies effectively.

\subsection{Diffusion-based Methods}

\subsubsection{Denoising Diffusion Probabilistic Models}
\label{sec:ddpm}

Denoising Diffusion Probabilistic Models (DDPMs)\cite{sohl2015deep, ho2020denoising} represent a significant advancement in generative modeling, particularly for high-dimensional data such as images. These models leverage a pair of complementary Markov processes: a forward diffusion process and a reverse denoising process. The core idea is to transform the original data into a noisy representation and then learn to reverse this process to generate new samples.

Let the original data be denoted as $\mathbf{x}_0$. The forward diffusion process incrementally adds noise to $\mathbf{x}_0$ over a series of steps, resulting in a sequence of increasingly noisy representations $\mathbf{x}_1, \mathbf{x}_2, \ldots, \mathbf{x}_T$. Each step is governed by a transition probability:
\begin{equation}
	q(\mathbf{x}_t | \mathbf{x}_{t-1}) = \mathcal{N}(\mathbf{x}_t; \sqrt{\alpha_t} \mathbf{x}_{t-1}, (1 - \alpha_t) \mathbf{I}),
\end{equation}
where $\alpha_t \in (0,1)$ controls the amount of noise added at each step, and $\mathcal{N}(\mathbf{x}; \boldsymbol{\mu}, \boldsymbol{\Sigma})$ represents a Gaussian distribution with mean $\boldsymbol{\mu}$ and covariance matrix $\boldsymbol{\Sigma}$. The cumulative effect of these steps can be described by:
\begin{equation}
	q(\mathbf{x}_t | \mathbf{x}_0) = \mathcal{N}(\mathbf{x}_t; \sqrt{\bar{\alpha}_t} \mathbf{x}_0, (1 - \bar{\alpha}_t) \mathbf{I}),
\end{equation}
where $\bar{\alpha}_t = \prod_{i=1}^t (1 - \alpha_i)$. This allows direct sampling from $\mathbf{x}_0$ to $\mathbf{x}_t$:
\begin{equation}
	\mathbf{x}_t = \sqrt{\bar{\alpha}_t} \mathbf{x}_0 + \sqrt{1 - \bar{\alpha}_t} \boldsymbol{\epsilon}, \quad \text{with} \quad \boldsymbol{\epsilon} \sim \mathcal{N}(\mathbf{0}, \mathbf{I}).
	\label{eq:forward_process}
\end{equation}
Typically, $\bar{\alpha}_T \approx 0$ is chosen to ensure that the final noisy representation $\mathbf{x}_T$ is effectively a standard Gaussian distribution.

The reverse denoising process aims to undo the noise added by the forward process. This is modeled using a parameterized neural network with parameters $\boldsymbol{\theta}$:
\begin{equation}
	p_{\boldsymbol{\theta}}(\mathbf{x}_{t-1} | \mathbf{x}_t) = \mathcal{N}(\mathbf{x}_{t-1}; \boldsymbol{\mu}_{\boldsymbol{\theta}}(\mathbf{x}_t, t), \boldsymbol{\Sigma}_{\boldsymbol{\theta}}(\mathbf{x}_t, t)).
\end{equation}

The goal is to learn the parameters $\boldsymbol{\theta}$ by minimizing the negative log-likelihood of the training data $\mathbf{x}_0$. This is achieved by minimizing a variational lower bound on the negative log-likelihood:
\begin{align}
	\boldsymbol{\theta}^* &= \arg\min_{\boldsymbol{\theta}} \mathbb{E}_{q(\mathbf{x}_{0:T})} \left[ -\log p(\mathbf{x}_T) - \sum_{t=1}^T \log \frac{p_{\boldsymbol{\theta}}(\mathbf{x}_{t-1} | \mathbf{x}_t)}{q(\mathbf{x}_t | \mathbf{x}_{t-1})} \right] \\
	&= \arg\min_{\boldsymbol{\theta}} \mathbb{E}_{q(\mathbf{x}_{0:T})} \left[ \sum_{t=2}^T D_{KL}(q(\mathbf{x}_{t-1} | \mathbf{x}_t, \mathbf{x}_0) \| p_\theta(\mathbf{x}_{t-1} | \mathbf{x}_t)) - \log p_\theta(\mathbf{x}_0 | \mathbf{x}_1) \right]. \label{eq:objective}
\end{align}

In order to effectively learn the reverse process by approximating the true posterior distribution $q(\mathbf{x}_{t-1}|\mathbf{x}_t, \mathbf{x}_0)$, DDPMs\cite{ho2020denoising} define:
\begin{align*}
	\boldsymbol{\mu}_{\boldsymbol{\theta}}(\mathbf{x}_t, t) &= \frac{1}{\sqrt{\alpha_t}} \left( \mathbf{x}_t - \frac{1 - \alpha_t}{\sqrt{1 - \bar{\alpha}_t}} \boldsymbol{\epsilon}_{\boldsymbol{\theta}}(\mathbf{x}_t, t) \right), \\
	\boldsymbol{\Sigma}_{\boldsymbol{\theta}}(\mathbf{x}_t, t) &= \sigma_t^2 \mathbf{I},
\end{align*}
where $\boldsymbol{\epsilon}_{\boldsymbol{\theta}}$ is a neural network that predicts the noise term corresponding to the input $\mathbf{x}_t$ and step $t$, and $\sigma_t^2 = \frac{(1 - \alpha_t)(1 - \bar{\alpha}_{t-1})}{1 - \bar{\alpha}_t}$.

By computing the Kullback-Leibler divergence, the objective function \eqref{eq:objective} can be further simplified to:
\begin{equation}
	\mathbb{E}_{t, \mathbf{x}_0, \boldsymbol{\epsilon}} \left[ \lambda(t) \left\| \boldsymbol{\epsilon} - \boldsymbol{\epsilon}_{\boldsymbol{\theta}} \left( \sqrt{\bar{\alpha}_t} \mathbf{x}_0 + \sqrt{1 - \bar{\alpha}_t} \boldsymbol{\epsilon}, t \right) \right\|^2 \right],
\end{equation}
where $\lambda(t) = \frac{\alpha_t^2}{2 \sigma_t^2 (1 - \alpha_t) (1 - \bar{\alpha}_t)}$ is a positive weight that can be ignored in practice for better performance.

To generate new samples, the reverse denoising process starts from $\mathbf{x}_T \sim \mathcal{N}(\mathbf{x}_T; \mathbf{0}, \mathbf{I})$ and iteratively removes noise. Specifically, for $t = T, T-1, \ldots, 1$:
\begin{equation}
	\mathbf{x}_{t-1} \gets \frac{1}{\sqrt{\alpha_t}} \left( \mathbf{x}_t - \frac{1 - \alpha_t}{\sqrt{1 - \bar{\alpha}_t}} \boldsymbol{\epsilon}_{\boldsymbol{\theta}}(\mathbf{x}_t, t) \right) + \sigma_t \mathbf{z},
\end{equation}
where $\mathbf{z} \sim \mathcal{N}(\mathbf{0}, \mathbf{I})$ for $t = T, T-1, \ldots, 2$, and $\mathbf{z} = \mathbf{0}$ for $t = 1$.

\subsubsection{Anomaly Detection}

With the rise of diffusion models, numerous studies have found that these models hold significant potential in image anomaly detection, particularly in medical image. For instance, Wolleb et al.\cite{wolleb2022diffusion} combines classifier guidance to achieve image-to-image translation between diseased and healthy subjects. AnoDDPM \cite{wyatt2022anoddpm} develops a multi-scale simplex noise diffusion process to address the issue of traditional Gaussian diffusion failing to capture larger anomalies.

Recently, some efforts have been made to extend diffusion models to multivariate time series anomaly detection. DiffusionAE \cite{pintilie2023time} integrates diffusion models with the reconstruction of AutoEncoder, enhancing the performance of anomaly detection. TimeADDM \cite{hu2024unsupervised} proposes encoding the original data into latent variables and then using diffusion models to reconstruct these latent variables.

It is important to note that the task of anomaly detection differs from generative tasks. The goal is to detect anomalies based on the differences between the original and reconstructed data, rather than generating diverse and realistic data. Therefore, the inference phase differs from the sampling process in DDPMs. Specifically, we need to apply the forward diffusion process to add noise to the test data and then use the reverse process to denoise it. Different diffusion steps can have varying impacts on anomaly detection. Even disregarding the quality of reconstruction, a small diffusion steps would render most of the training ineffective, whereas a large diffusion steps can significantly degrade the efficiency of reconstruction. On this issue, current works have adopted differing strategies. Wolleb et al.\cite{wolleb2022diffusion} uses complete diffusion, while AnoDDPM\cite{wyatt2022anoddpm} and DiffusionAE\cite{pintilie2023time} adopt partial diffusion. TimeADDM\cite{hu2024unsupervised} flexibly uses a weighted sum of reconstructions from different diffusion steps. However, determining the optimal diffusion steps often requires extensive experimentation.

To address these limitations, this paper proposes a method that combines the denoiser of diffusion model with generative adversarial network. This approach simultaneously generates noisy data and predicts the denoising steps required, providing a flexible solution to the aforementioned limitations.

\section{Methods}

\subsection{Problem Formulation}

In the context of multivariate time series anomaly detection (MTS-AD), the primary objective is to identify unusual patterns or outliers in the time series data. Specifically, we aim to classify each time point in a multivariate time series as either normal or anomalous.

Formally, let $ S = [s_1, s_2, \ldots, s_T]^T \in \mathbb{R}^{T \times D} $ represent a collection of $ D $-dimensional multivariate time series, where $ s_t \in \mathbb{R}^D $ denotes the observation vector at the $ t $-th time point. Each $ s_t $ consists of $ D $ features measured at the same time point. The goal is to assign a binary label $ y_t \in \{0, 1\} $ to each $ s_t $, where $ y_t = 1 $ indicates that $ s_t $ is an anomaly, and $ y_t = 0 $ indicates that $ s_t $ is a normal observation.

\subsection{Preprocessing of MTS}
\label{sec:preprocess}
To ensure the effectiveness and efficiency of anomaly detection in multivariate time series, proper preprocessing is essential. This section outlines key preprocessing steps, including data normalization and time series segmentation.

\subsubsection{Normalization}
To enhance the robustness and generalization of the model, data normalization is performed to standardize the scale of different features. The normalization process scales each feature to a common range, typically $[0, 1]$. Formally, the normalization formula is as follows:
\begin{equation}
	S_{\text{norm}}[t, d] = \frac{S[t, d] - \displaystyle\min_{1 \leq j \leq T} S[j, d]}{\displaystyle\max_{1 \leq j \leq T} S[j, d] - \displaystyle\min_{1 \leq j \leq T} S[j, d]}, \quad d = 1, \ldots, D.
\end{equation}

This transformation ensures that all features contribute equally to the model's learning process, thereby improving the model's performance.

\subsubsection{Sliding Window Technique}
To effectively capture the temporal dependencies and patterns in MTS, the data is often processed using a sliding window technique\cite{w13131862}. This technique divides the time series into smaller, manageable chunks, facilitating the application of machine learning algorithms, especially those that require fixed-size inputs, such as neural networks. Therefore, we need to apply the sliding window technique to the training data.

Formally, we set a sliding window size $ w $ and a sliding step $ l $, the time series $ S = [s_1, s_2, \ldots, s_T]^T \in \mathbb{R}^{T \times D} $ is then divided into segments as follows:

$$
W_1 = [s_1, s_2, \ldots, s_w]^T,
$$
$$
W_2 = [s_{1+l}, s_{1+l+1}, \ldots, s_{1+l+(w-1)}]^T,
$$
$$
\vdots
$$
$$
W_k = [s_{1+(k-1)l}, s_{1+(k-1)l+1}, \ldots, s_{1+(k-1)l+(w-1)}]^T,
$$
where $ k = \left\lfloor \frac{T - w}{l} \right\rfloor + 1 $ is the total number of windows, and $\left\lfloor \cdot \right\rfloor$ denotes the floor function.

The benefits of the sliding window technique include:
\begin{itemize}
	\item Sliding windows help in capturing the temporal dependencies inherent in time series data, which is crucial for identifying patterns and anomalies.
	\item Diffusion models require large amounts of data for training. The sliding window technique increases the amount of training data, letting the model training be more thorough.
\end{itemize}

\subsection{Architecture of DiffGAN}

\begin{figure}[H]
	\centering
	\includegraphics[width=1.0\textwidth, height=0.4\textheight]{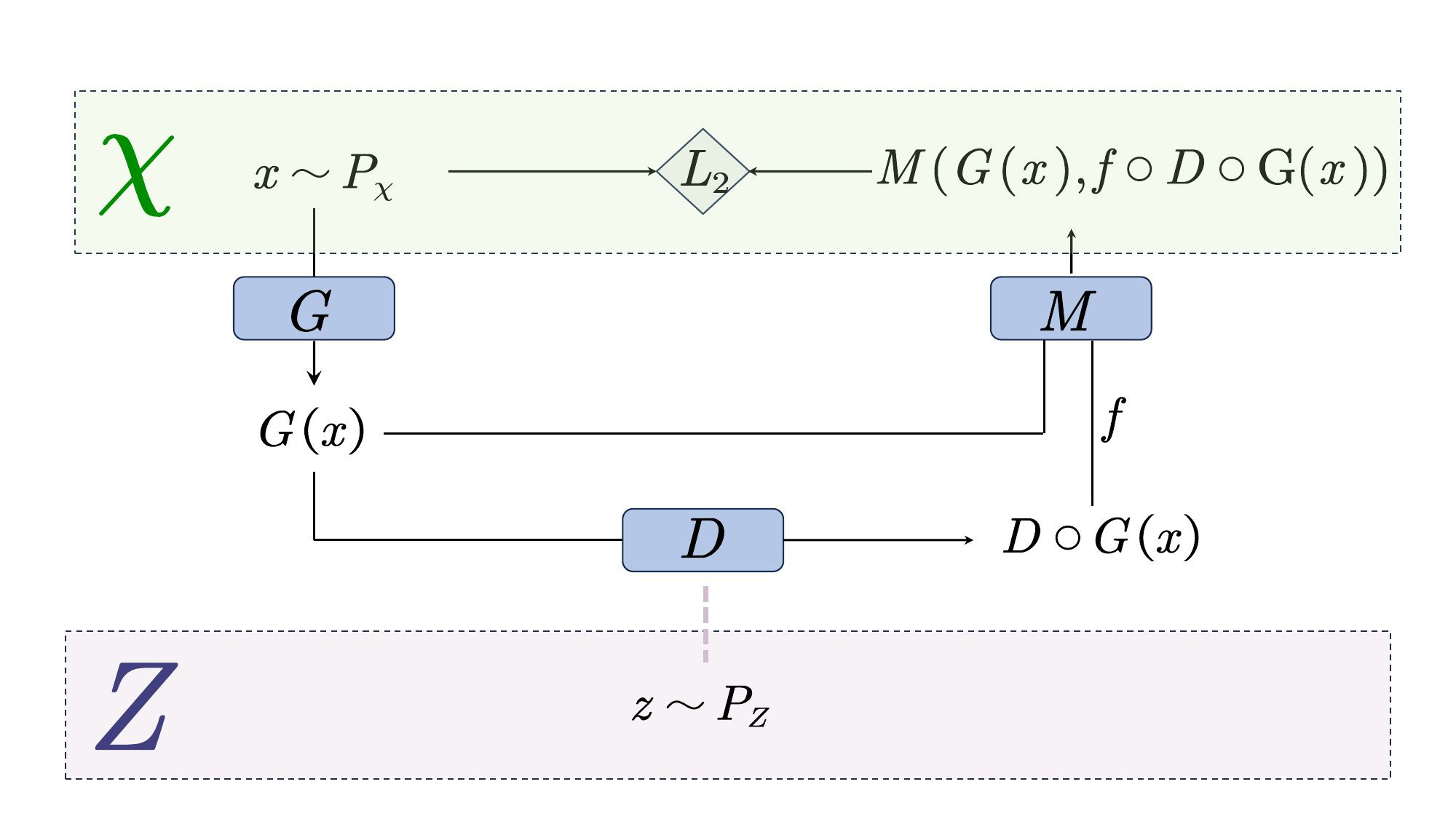}
	\caption{Architecture of DiffGAN.}
	\label{fig:framework}
\end{figure}

The main aim of our method is to design a GAN compatible with the denoiser of diffusion model, enabling the use of the generator to replace the forward process. Figure~\ref{fig:framework} shows the overall framework of DiffGAN, where M denotes the denoising module with denoiser, G the generator, and D the discriminator. The entire process is divided into three stages: training the denoiser, training the generator and discriminator, and performing anomaly detection. The detailed content will be elaborated in the following three sections.

\subsection{Training the Denoiser}
Similar to most works on diffusion models, the architecture of our denoiser network adopts the classic U-Net structure \cite{ronneberger2015u}. The U-Net employs an encoder-decoder architecture where the encoder gradually reduces spatial resolution to extract high-level features, while the decoder increases spatial resolution to restore details. This design is particularly suitable for the task of progressively removing noise from data and reconstructing the original signal.

For such a complex neural network architecture as U-Net, we utilize AdamW\cite{loshchilov2017decoupled} as the optimization method. AdamW combines the advantages of the Adam optimizer with weight decay regularization, typically achieving better convergence properties compared to SGD for architectures like U-Net.

The training principle of the denoiser is detailed in Section~\ref{sec:ddpm}. Let $ S = [s_1, s_2, \ldots, s_T]^T \in \mathbb{R}^{T \times D} $ represent the multivariate time series used for training, we normlize and apply sliding window technique to $S$, resulting in $\mathcal{W} = \{W_1, W_2, \ldots, W_k\}$, using the same notation as previously mentioned. The training pseudocode is outlined in Algorithm~\ref{alg:training_denoiser}.

Based on the denoiser $\hat{\mathcal{E}} = \mathcal{E}_{\theta^*}$ trained via Algorithm~\ref{alg:training_denoiser}, we define the denoising module $M$ as follows:
\begin{equation}
	M(\mathbf{x}_n, n) = M_1 \circ M_{2} \circ \cdots \circ M_n (\mathbf{x}_n),
\end{equation}
where each $M_n$ is defined by:
\begin{equation}
	M_n(\mathbf{x}_n) =
	\begin{cases} 
		\frac{1}{\sqrt{\alpha_n}} \left( \mathbf{x}_n - \frac{1 - \alpha_n}{\sqrt{1 - \bar{\alpha}_n}} \hat{\mathcal{E}}(\mathbf{x}_n, n) \right) + \sigma_n \mathbf{z}, & \text{for } n = 2, 3, \ldots, N ;\\
		\frac{1}{\sqrt{\alpha_1}} \left( \mathbf{x}_1 - \frac{1 - \alpha_1}{\sqrt{1 - \bar{\alpha}_1}} \hat{\mathcal{E}}(\mathbf{x}_1, 1) \right), & \text{for } n = 1.
	\end{cases}
	\label{eq:m_n}
\end{equation}
with $\sigma_n = \sqrt{\frac{(1 - \alpha_n)(1 - \bar{\alpha}_{n-1})}{1 - \bar{\alpha}_n}}$ and $\mathbf{z} \sim \mathcal{N}(\mathbf{0}, \mathbf{I})$.

\begin{algorithm}[H]
	\caption{Training the Denoiser}\label{alg:training_denoiser}
	\begin{algorithmic}[1]
		\Require $ \mathcal{W} $: the set of sliding windows used for training.
		\Statex \hspace*{\algorithmicindent} $ \mathcal{E}_\theta $: untrained denoiser.
		\Statex \hspace*{\algorithmicindent} $ N \in \mathbb{N}^+ $: total diffusion steps.
		\Statex \hspace*{\algorithmicindent} $ m \in \mathbb{N}^+ $: batch size.
		\Statex \hspace*{\algorithmicindent} $ \alpha \in \mathbb{R}^+ $: learning rate.
		\Repeat
		\State Sample a batch $\{W_1, W_2, \ldots, W_m\}$ from $\mathcal{W}$.
		\State Sample $ n \sim \text{Uniform}\{1, 2, \ldots, N\} $ and $ \epsilon \sim \mathcal{N}(\mathbf{0}, \mathbf{I}) $.
		\State $ g_\theta \leftarrow \nabla_\theta \left[ \frac{1}{m} \sum_{i=1}^{m} \left\| \epsilon - \mathcal{E}_\theta(\sqrt{\bar{\alpha}_n} W_i + \sqrt{1-\bar{\alpha}_n} \epsilon, n) \right\|_2^2 \right] $.
		\State $ \theta \leftarrow \theta - \alpha \cdot \text{AdamW}(\theta, g_\theta) $.
		\Until{converged}
	\end{algorithmic}
	\noindent \textbf{Output:} $ \mathcal{E}_{\theta^*} $: trained denoiser.
\end{algorithm}

\subsection{Training the Generator and Discriminator}

The forward process of diffusion models is essentially a step-by-step noise addition process based on a Markov chain. It gradually transforms the original data into pure noise through a series of discrete time steps. Indeed, as indicated by Equation \eqref{eq:forward_process}, the forward process at each time step can be represented as single-step noise addition, with the noise intensity being entirely dependent on the current time step. To design a reconstruction model based on partial diffusion, we initially do not know what level of noise intensity would enhance the model's performance. Therefore, we consider leveraging other models to more flexibly control and measure the noise addition process. Since the forward process can be viewed as single-step noise addition, this process can also be implemented using the generator of GAN. To guide the generator in transforming the original data towards pure noise, we feed the generator's output along with pure noise into the discriminator, leading to the following adversarial loss:
\begin{equation}
	\min_G \max_D \mathbb{E}_{z \sim P_z}[ \log D(z) ] + \mathbb{E}_{x \sim P_{\text{data}}}[ \log (1 - D(G(x))) ].
\end{equation}

Unlike traditional GANs, the generator and discriminator in this context serve to construct a reconstruction model based on partial diffusion rather than generating realistic pure noise. Therefore, we need to re-examine the role of the discriminator. The output of the discriminator can be interpreted as the probability that the generated data is pure noise, with this probability reflecting the degree of similarity between the generated data and pure noise. This implies that the discriminator's output can be aligned with the time steps in the forward process. To achieve this, we design a function $f$ that maps the probability to the time step, for instance,
\begin{align*}
	f: [0, 1] &\to \{0, 1, \ldots, N\} \\
	p &\mapsto \left[ N \cdot p \right],
\end{align*}
Or, based on the noise intensity at each time step, $f$ can also be defined as:
\begin{align*}
	f: [0, 1] &\to \{0, 1, \ldots, N\} \\
	p &\mapsto \max \{ n \mid 1 - \bar{\alpha_n} \leq p \},
\end{align*}
Here, $N$ represents the total diffusion steps.

Up to this point, the GAN simultaneously provides the noised data and the corresponding time steps required by the reconstruction model. We feed both of these into the denoiser and perform the denoising module. 

The quality of reconstruction is crucial for enhancing the effectiveness of anomaly detection. Therefore, we need to incorporate a reconstruction loss during the training process of the GAN. This enables the GAN to leverage the inherent features of the original data, thereby controlling noise levels that facilitate effective reconstruction. The reconstruction loss is defined as follows:
\begin{equation}
	\min_{G, D} \mathbb{E}_{x \sim P_{\text{data}}} \left\| x - M(G(x), f \circ D \circ G(x)) \right\|_2^2.
\end{equation}

In summary, we summarize the training pseudocode for the generator and discriminator as Algorithm~\ref{alg:training_g_d}.
\begin{algorithm}[H]
	\caption{Training the Generator and Discriminator}\label{alg:training_g_d}
	\begin{algorithmic}[1]
		\Require $ \mathcal{W} $: the set of sliding windows used for training.
		\Statex \hspace*{\algorithmicindent} $ G_\theta $: untrained generator.
		\Statex \hspace*{\algorithmicindent} $ D_\varphi $: untrained discriminator.
		\Statex \hspace*{\algorithmicindent} $ M $: denoising module with trained denoiser.
		\Statex \hspace*{\algorithmicindent} $ \lambda \in \mathbb{R}^+ $: weight of adversarial loss.
		\Statex \hspace*{\algorithmicindent} $ m \in \mathbb{N}^+ $: batch size.
		\Statex \hspace*{\algorithmicindent} $ \alpha_g, \alpha_d \in \mathbb{R}^+$: learning rate.
		\Repeat
		\State Sample a batch $\{W_1, W_2, \ldots, W_m\}$ from $\mathcal{W}$.
		\State Sample $z \sim \mathcal{N}(\mathbf{0}, \mathbf{I})$.
		\State $g_\varphi \leftarrow \nabla_\varphi \left\{ \frac{1}{m} \sum_{i=1}^m \left[ \lambda \left( \log D_\varphi \circ G_\theta(W_i) + \log (1 - D_\varphi(z)) \right) \right. \right.$
		\State $\quad \quad \quad \quad \quad \left. + \| W_i - M(G_\theta(W_i), f \circ D_\varphi \circ G_\theta(W_i)) \|_2^2 \right] \left. \right\}$.
		\State $\varphi \leftarrow \varphi - \alpha_d \cdot \text{AdamW}(\varphi, g_\varphi)$.
		\State $g_\theta \leftarrow \nabla_\theta \left\{ \frac{1}{m} \sum_{i=1}^m \left[ \lambda \left( \log D_\varphi(z) + \log (1 - D_\varphi \circ G_\theta(W_i)) \right) \right. \right.$
		\State $\quad \quad \quad \quad \quad \left. + \| W_i - M(G_\theta(W_i), f \circ D_\varphi \circ G_\theta(W_i)) \|_2^2 \right] \left. \right\}$.
		\State $\theta \leftarrow \theta - \alpha_g \cdot \text{AdamW}(\theta, g_\theta)$.
		\Until{converged}
	\end{algorithmic}
	\noindent \textbf{Output:} $ G_{\theta^*}, D_{\varphi^*} $: trained generator and discriminator.
\end{algorithm}

\subsection{Anomaly detection based on DiffGAN}

After all modules of DiffGAN have been fully trained, the model has learned the typical characteristics and distribution patterns of normal data. Normal data usually exhibit lower reconstruction error, whereas anomalous data tend to have higher reconstruction error. Therefore, we define the anomaly score as the reconstruction error:
\begin{equation}
	e(x) = \| x - M(G(x), f \circ D \circ G(x)) \|_2^2.
\end{equation}

We summarize the pseudocode for anomaly detection as Algorithm~\ref{alg:anomaly_detection}.
\begin{algorithm}[H]
	\caption{Anomaly detection based on DiffGAN}\label{alg:anomaly_detection}
	\begin{algorithmic}[1]
		\Require $ S \in \mathbb{R}^{T \times D} $: $ D $-dimensional multivariate time series used for testing.
		\Statex \hspace*{\algorithmicindent} $ G, D, M $: trained modules of DiffGAN.
		\Statex \hspace*{\algorithmicindent} $ \delta \in \mathbb{R}^+ $: threshold of the anomaly score.
		\State $ \hat{S} \leftarrow M(G(S), f \circ D \circ G(S)) $.
		\State $ e_t \leftarrow \| S[t,:] - \hat{S}[t,:] \|_2^2 $ for $ t = 1, 2, \ldots, T $.
		\State $ \hat{y}_t \leftarrow \epsilon(e_t - \delta) $ for $ t = 1, 2, \ldots, T $. \quad \quad \quad \quad \quad$\triangleright$ $\epsilon(\cdot)$ denotes step function.
	\end{algorithmic}
	\noindent \textbf{Output:} $ \hat{y} = [\hat{y_1}, \hat{y_2}, \ldots, \hat{y_T}] $: predicted binary labels.
\end{algorithm}

\subsection{Further Explorations}

While the proposed DiffGAN model exhibits promising performance, it inevitably inherits certain drawbacks from DDPMs, such as low reconstruction efficiency. To address these issues, we have conducted further explorations.

One attempt involved designing the denoising module $ M $ for single-step denoising. For instance, keeping the denoiser unchanged, we redefined the denoising module $ M $ as follows:
\begin{equation}
	M(x_n, n) = \frac{x_n - \sqrt{1 - \bar{\alpha}_n} \hat{\mathcal{E}}(x_n, n)}{\sqrt{\bar{\alpha}_n}}.
\end{equation}

Additionally, we sought to optimize the denoising module $ M $ by leveraging improved algorithms from the field of image generation. For example, we applied the sampling algorithm from Denoising Diffusion Implicit Models (DDIMs)\cite{song2020denoising}, modifying Equation (\ref{eq:m_n}) to:
\begin{align*}
	M_n(x_n) &= \sqrt{\bar{\alpha}_{n-1}} \cdot \frac{x_n - \sqrt{1 - \bar{\alpha}_n} \hat{\mathcal{E}}(x_n, n)}{\sqrt{\bar{\alpha}_n}} + \sqrt{1 - \bar{\alpha}_{n-1}} \hat{\mathcal{E}}(x_n, n) \\
	&= \frac{x_n - \sqrt{1 - \bar{\alpha}_n} \hat{\mathcal{E}}(x_n, n)}{\sqrt{\alpha}_n} + \sqrt{1 - \bar{\alpha}_{n-1}} \hat{\mathcal{E}}(x_n, n).
\end{align*}

However, while these attempts improved model efficiency, they did not achieve superior performance in anomaly detection. Consequently, they were not adopted in this paper. This indicates that there remains significant scope for further research in this domain.

\section{Experiments}

\subsection{Datasets}

To evaluate the performance of the proposed DiffGAN, we carry out experiments on five multivariate time series datasets: Global Point, Contextual Point, Seasonal Pattern, Shapelet Pattern, Trend Pattern. These five datasets were created by ~\cite{pintilie2023time}, encompassing all types of anomalies in time series as proposed by NeurIPS-TS~\cite{lai2021revisiting}. Each dataset contains 50,000 timesteps and 5 dimensions, with one dimension containing anomalies. The datasets are divided into training, validation, and testing sets in a ratio of 2:1:2. Experiments commence after the preprocessing detailed in Section~\ref{sec:preprocess}.

\subsection{Baselines}

In this section, we compare our proposed DiffGAN model with several baseline models. The motivation for proposing the DiffGAN model stems from the limitations of traditional diffusion models, so we first compare it with diffusion models that use different numbers of diffusion steps. Since almost all novel MTS anomaly detection models have already been compared with classical models as described in Section~\ref{sec:classical} and have achieved superior results, here we select two state-of-the-art reconstruction models as baseline models: TadGAN(based on GANs) and DiffusionAE(based on DMs).

\textbf{Diffusion}\cite{ho2020denoising, pintilie2023time}: The denoiser is designed as a U-Net. During training of the denoiser, the total diffusion steps $N=100$, while during anomaly detection, the diffusion steps is set to $M=\{20, 50, 80\}$, corresponding to models named Diffusion-20, Diffusion-50, and Diffusion-80 respectively.

\textbf{TadGAN}\cite{geiger2020tadgan}: The encoder and decoder are designed as LSTMs, and the two critics are designed as MLPs.

\textbf{DiffusionAE}\cite{pintilie2023time}: Following the authors' experiments, the AutoEncoder is designed as a Transformer, and the denoiser is designed as a U-Net. During training of the denoiser, the total diffusion steps $N=100$. For anomaly detection, the diffusion steps is chosen from $M=\{20, 50, 80\}$ to achieve better performance.

\textbf{DiffGAN}: The denoiser is designed as a U-Net, the generator as an LSTM, and the discriminator as an MLP. During training of the denoiser, the total diffusion steps $N=100$. While training the generator and discriminator, the weight coefficient for adversarial loss is set to $\lambda=0.7$.

\subsection{Metrics}

Anomaly detection in multivariate time series is a binary classification task. Among the evaluation metrics used, Precision (P), Recall (R), and F1 Score are particularly important due to their ability to provide a comprehensive assessment of a model's accuracy in predicting anomalies.

\textbf{Precision (P)} measures the proportion of true positive predictions among all positive predictions made by the model. It indicates how many of the detected anomalies are actual anomalies. Mathematically, precision is defined as:
\begin{equation}
	P = \frac{\text{True Positives (TP)}}{\text{True Positives (TP)} + \text{False Positives (FP)}},
\end{equation}
where True Positives (\textit{TP}) are the correctly identified anomalies, and False Positives (\textit{FP}) are the non-anomalous instances incorrectly labeled as anomalies. A higher precision means that the model has fewer false alarms.

\textbf{Recall (R)}, also known as Sensitivity or True Positive Rate, measures the proportion of true positive predictions among all actual anomalies present in the dataset. It tells us how many of the actual anomalies were caught by the model. The formula for recall is given by:
\begin{equation}
	R = \frac{\text{True Positives (TP)}}{\text{True Positives (TP)} + \text{False Negatives (FN)}},
\end{equation}
where False Negatives (\textit{FN}) refer to the anomalies that were missed by the model. High recall signifies that the model successfully captures most of the anomalies.

\textbf{F1 Score (F1)} is the harmonic mean of precision and recall, providing a single metric that balances both. It is especially useful when the class distribution is uneven or when you seek a balance between precision and recall. The F1 Score is calculated using the following equation:
\begin{equation}
	F1 = 2 \cdot \frac{P \cdot R}{P + R},
\end{equation}
This metric reaches its best value at 1 and worst score at 0. An F1 score of 1 implies perfect precision and recall, indicating no false positives or negatives.

These metrics together offer a comprehensive assessment of an anomaly detection model's effectiveness in distinguishing between normal and abnormal points in multivariate time series.

\subsection{Results}

We evaluated the performance of various benchmark models and our proposed DiffGAN on five multivariate time series datasets, as detailed in Table \ref{tab:performance}. It is important to note that the point-adjustment evaluation strategy can enhance metrics on some datasets\cite{kim2022towards}, but this does not significantly influence model comparisons. Therefore, this strategy was not adopted in this experiment.

Although our proposed DiffGAN did not achieve the highest Precision (P) or Recall (R), it outperformed all benchmark models in terms of the F1 score on multiple datasets. This highlights its significant potential for multivariate time series anomaly detection.

\begin{figure}[H]
	\centering
	\includegraphics[width=1.0\textwidth, height=0.4\textheight]{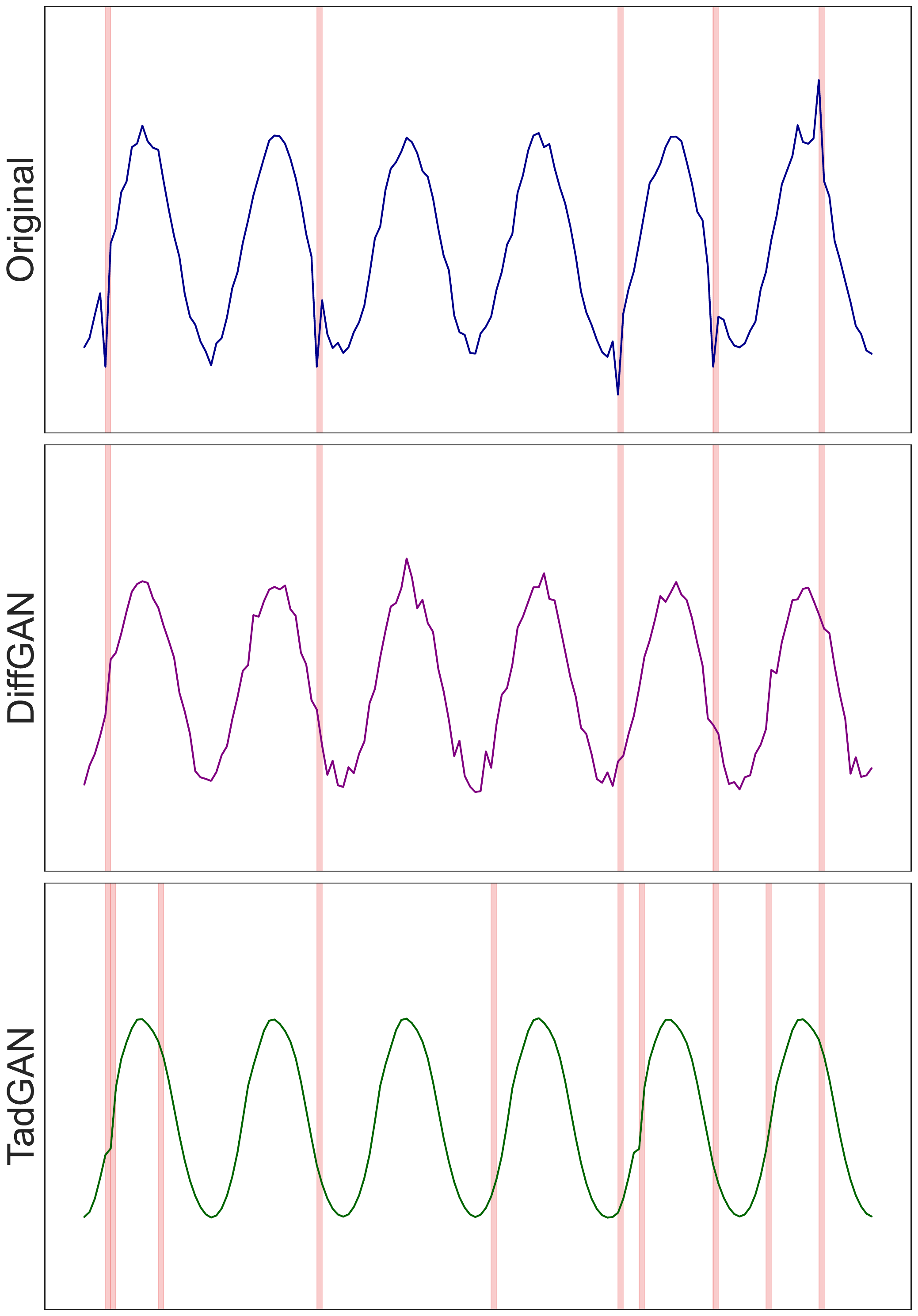}
	\caption{The reconstruction and anomaly detection outcomes of TadGAN and DiffGAN on a segment of the Global dataset.}
	\label{fig:recon}
\end{figure}

\begin{figure}[H]
	\centering
	\includegraphics[width=1.0\textwidth, height=0.4\textheight]{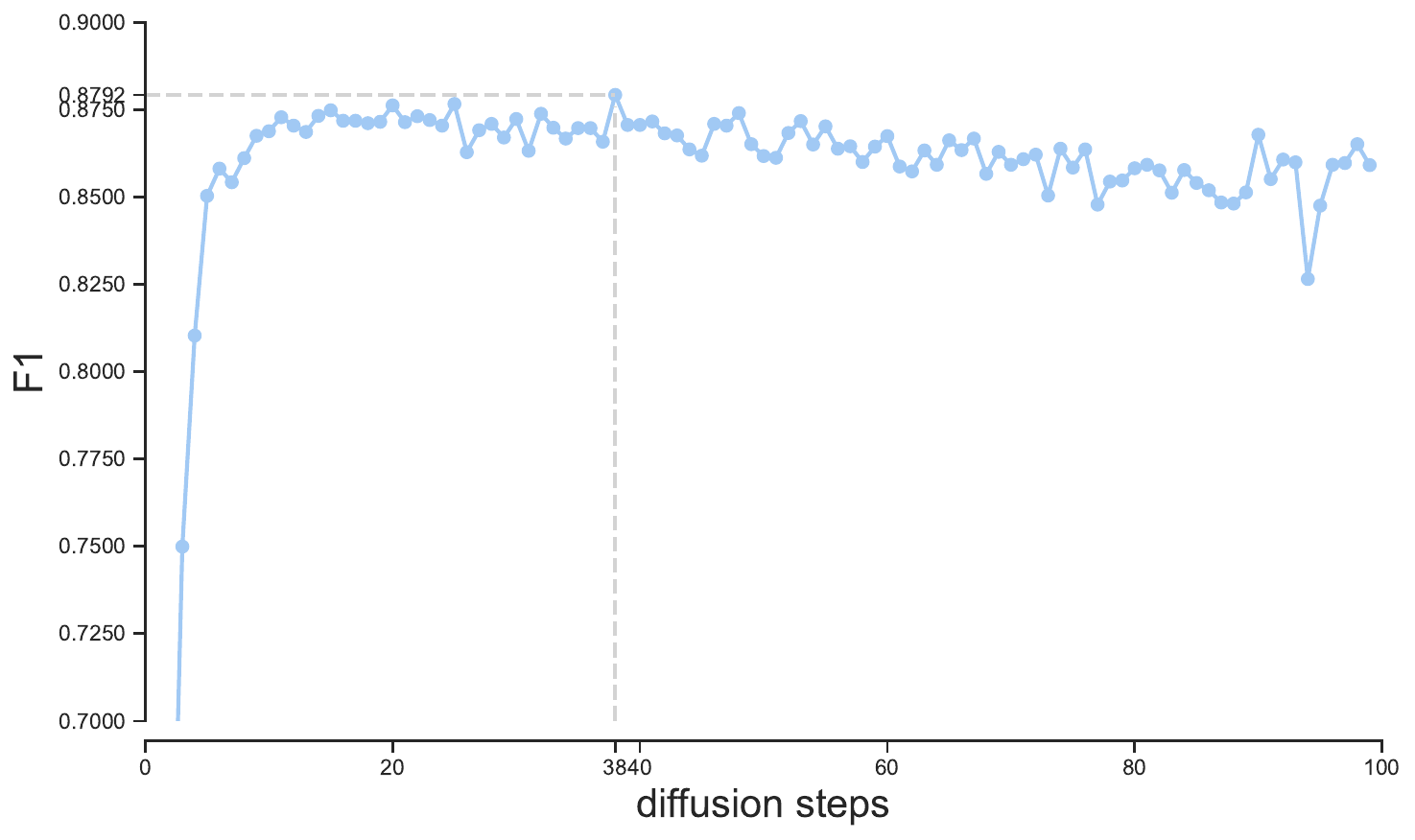}
	\caption{F1 score varies with the number of diffusion steps using diffusion model on the Global dataset.}
	\label{fig:f1}
\end{figure}

Furthermore, from Table \ref{tab:performance}, we observed that:

\begin{itemize}
	\item The GAN-based model TadGAN demonstrates overall weaker performance compared to diffusion-based models. Figure \ref{fig:recon} illustrates the reconstruction and anomaly detection outcomes of TadGAN and DiffGAN on a segment of the Global dataset. TadGAN's reconstruction process involves a complete mapping from latent space to data space, resulting in smoother sequences that fail to accurately capture true mutations within the sequence, thus causing misclassification. In contrast, DiffGAN employs a partial diffusion strategy, offering superior reconstruction quality and reducing the likelihood of misclassification.
	\item Regarding the impact of diffusion steps on diffusion model, one might assume that fewer diffusion steps yield better results. However, this is not necessarily the case. We found that when the number of diffusion steps $M < 5$, the model cannot perform effective anomaly detection. Figure \ref{fig:f1} presents a line graph showing how the F1 score varies with the number of diffusion steps using diffusion model on the Global dataset. Considering both efficiency and performance in anomaly detection, diffusion model with moderately small diffusion steps are preferable. If the focus is solely on maximizing anomaly detection performance, our proposed DiffGAN model stands out, as indicated by Figure \ref{fig:f1} and Table \ref{tab:performance}, the F1 scores of DiffGAN surpass those of any diffusion model across all diffusion step counts.
\end{itemize}

\begin{table}[H]
	\centering
	\footnotesize
	\setlength\tabcolsep{3pt}
	\begin{tabular}{lcccccccccc}
		\toprule
		& \multicolumn{3}{c}{Diffusion-20} & \multicolumn{3}{c}{Diffusion-50} & \multicolumn{3}{c}{Diffusion-80} \\
		\cmidrule(lr){2-4} \cmidrule(lr){5-7} \cmidrule(lr){8-10}
		& P & R & F1 & P & R & F1 & P & R & F1 \\
		\midrule
		Global & \textbf{0.9605} & 0.8056 & 0.8762 & 0.9321 & 0.8012 & 0.8617 & 0.9483 & 0.7838 & 0.8582 \\
		Contextual & \textbf{0.9194} & 0.5963 & 0.7234 & 0.8841 & 0.6251 & 0.7324 & 0.8936 & 0.6077 & 0.7234 \\
		Seasonal & 0.9629 & 0.7098 & 0.8172 & \textbf{0.9768} & 0.7149 & 0.8256 & 0.9701 & 0.7194 & 0.8261 \\
		Shapelet & 0.6013 & 0.4055 & 0.4843 & \textbf{0.7608} & 0.3501 & 0.4795 & 0.7235 & 0.3730 & 0.4922 \\
		Trend & 0.7708 & \textbf{0.4984} & \textbf{0.6054} & \textbf{0.7943} & 0.4189 & 0.5485 & 0.2592 & 0.3845 & 0.3097 \\
		\midrule
		& \multicolumn{3}{c}{TadGAN} & \multicolumn{3}{c}{DiffusionAE} & \multicolumn{3}{c}{\textbf{DiffGAN}} \\
		\cmidrule(lr){2-4} \cmidrule(lr){5-7} \cmidrule(lr){8-10}
		& P & R & F1 & P & R & F1 & P & R & F1 \\
		\midrule
		Global & 0.8111 & 0.8422 & 0.8264 & 0.9532 & 0.8221 & 0.8828 & 0.9576 & \textbf{0.8274} & \textbf{0.8877} \\
		Contextual & 0.7203 & 0.6286 & 0.6713 & 0.8961 & \textbf{0.6492} & \textbf{0.7529} & 0.8985 & 0.6385 & 0.7465 \\
		Seasonal & 0.6849 & 0.6485 & 0.6662 & 0.9684 & 0.7265 & 0.8302 & 0.9692 & \textbf{0.7324} & \textbf{0.8343} \\
		Shapelet & 0.6067 & 0.3213 & 0.4201 & 0.7354 & 0.4167 & 0.5320 & 0.7265 & \textbf{0.4357} & \textbf{0.5447} \\
		Trend & 0.2765 & 0.4391 & 0.3394 & 0.7461 & 0.3827 & 0.5059 & 0.6842 & 0.4568 & 0.5478 \\
		\bottomrule
	\end{tabular}
	\caption{Performance comparison.}
	\label{tab:performance}
\end{table}

\section{Conclusion}
In this paper, to address the issue of partial diffusion uncertainty associated with diffusion models, we introduce a novel DiffGAN model for anomaly detection in multivariate time series. On one hand, leveraging the flexibility of generative adversarial network, we replace the forward diffusion process with the generator under the control of the discriminator. On the other hand, we utilize the denoiser from diffusion models to achieve superior reconstruction quality. Finally, experimental results demonstrate that DiffGAN exhibits superior performance in anomaly detection for multivariate time series.

\bibliographystyle{plain}
\bibliography{references.bib}

\begin{thebibliography}{10}

\bibitem{ali2017detecting}
Saqib Ali, Guojun Wang, Roger~Leslie Cottrell, and Tayyba Anwar.
\newblock Detecting anomalies from end-to-end internet performance measurements (pinger) using cluster-based local outlier factor.
\newblock In {\em 2017 IEEE International Symposium on Parallel and Distributed Processing with Applications and 2017 IEEE International Conference on Ubiquitous Computing and Communications (ISPA/IUCC)}, pages 982--989. IEEE, 2017.

\bibitem{geiger2020tadgan}
Alexander Geiger, Dongyu Liu, Sarah Alnegheimish, Alfredo Cuesta-Infante, and Kalyan Veeramachaneni.
\newblock Tadgan: Time series anomaly detection using generative adversarial networks.
\newblock In {\em 2020 IEEE International Conference on Big Data (Big Data)}, pages 33--43. IEEE, 2020.

\bibitem{goodfellow2014generative}
Ian Goodfellow, Jean Pouget-Abadie, Mehdi Mirza, Bing Xu, David Warde-Farley, Sherjil Ozair, Aaron Courville, and Yoshua Bengio.
\newblock Generative adversarial nets.
\newblock In {\em Advances in Neural Information Processing Systems}, volume~27, pages 2672--2680. NeurIPS, 2014.

\bibitem{guo2018knn}
Jinyu Guo, Xin Wang, and Yuan Li.
\newblock knn based on probability density for fault detection in multimodal processes.
\newblock {\em Journal of Chemometrics}, 32(7):e3021, 2018.

\bibitem{ho2020denoising}
Jonathan Ho, Ajay Jain, and Pieter Abbeel.
\newblock Denoising diffusion probabilistic models.
\newblock In {\em Advances in Neural Information Processing Systems}, volume~33, pages 6840--6851. NeurIPS, 2020.

\bibitem{hu2024unsupervised}
Rongyao Hu, Xinyu Yuan, Yan Qiao, Benchu Zhang, and Pei Zhao.
\newblock Unsupervised anomaly detection for multivariate time series using diffusion model.
\newblock In {\em 2024 IEEE International Conference on Acoustics, Speech and Signal Processing (ICASSP)}, pages 9606--9610. IEEE, 2024.

\bibitem{jain2022modified}
Praphula~Kumar Jain, Mani~Shankar Bajpai, and Rajendra Pamula.
\newblock A modified dbscan algorithm for anomaly detection in time-series data with seasonality.
\newblock {\em International Arab Journal of Information Technology}, 19(1):23--28, 2022.

\bibitem{kim2022towards}
Siwon Kim, Kukjin Choi, Hyun-Soo Choi, Byunghan Lee, and Sungroh Yoon.
\newblock Towards a rigorous evaluation of time-series anomaly detection.
\newblock In {\em Proceedings of the AAAI Conference on Artificial Intelligence}, volume~36, pages 9015--9024. AAAI Press, 2022.

\bibitem{w13131862}
Lattawit Kulanuwat, Chantana Chantrapornchai, Montri Maleewong, Papis Wongchaisuwat, Supaluk Wimala, Kanoksri Sarinnapakorn, and Surajate Boonya-aroonnet.
\newblock Anomaly detection using a sliding window technique and data imputation with machine learning for hydrological time series.
\newblock {\em Water}, 13(13):1862, 2021.

\bibitem{lai2021revisiting}
Kwei-Herng Lai, Daochen Zha, Junjie Xu, Yue Zhao, Guanchu Wang, and Xia Hu.
\newblock Revisiting time series outlier detection: Definitions and benchmarks.
\newblock In {\em Thirty-Fifth Conference on Neural Information Processing Systems (NeurIPS) Datasets and Benchmarks Track}. NeurIPS, 2021.

\bibitem{li2019mad}
Dan Li, Dacheng Chen, Baihong Jin, Lei Shi, Jonathan Goh, and See-Kiong Ng.
\newblock Mad-gan: Multivariate anomaly detection for time series data with generative adversarial networks.
\newblock In {\em International Conference on Artificial Neural Networks (ICANN)}, Lecture Notes in Computer Science, pages 703--716. Springer, 2019.

\bibitem{loshchilov2017decoupled}
Ilya Loshchilov and Frank Hutter.
\newblock Decoupled weight decay regularization.
\newblock {\em arXiv preprint arXiv:1711.05101}, 2017.

\bibitem{nakamura2021anomaly}
Takumi Nakamura, Keisuke Yamaguchi, Takayuki Sakamoto, Masahiro Kimura, and Yoshihiko Sako.
\newblock Anomaly detection for predictive maintenance of industrial machinery using multivariate time series data.
\newblock In {\em ICASSP, IEEE International Conference on Acoustics, Speech and Signal Processing - Proceedings}, pages 8658--8662. IEEE, 2021.

\bibitem{pintilie2023time}
Ioana Pintilie, Andrei Manolache, and Florin Brad.
\newblock Time series anomaly detection using diffusion-based models.
\newblock In {\em 2023 IEEE International Conference on Data Mining Workshops (ICDMW)}, pages 570--578. IEEE, 2023.

\bibitem{ren2020timeseries}
Hansheng Ren, Bixiong Xu, Jie Chen, Yuyang Wang, Alexander~D. Kent, Shobha Venkataraman, Zhen Qin, Yi~Wang, and Weiyi Ge.
\newblock Time-series anomaly detection service at microsoft.
\newblock {\em arXiv preprint arXiv:2009.05248}, 2020.

\bibitem{ronneberger2015u}
Olaf Ronneberger, Philipp Fischer, and Thomas Brox.
\newblock U-net: Convolutional networks for biomedical image segmentation.
\newblock In {\em Medical Image Computing and Computer-Assisted Intervention -- MICCAI 2015: 18th International Conference, Munich, Germany, October 5-9, 2015, Proceedings, Part III}, volume 9351 of {\em Lecture Notes in Computer Science}, pages 234--241. Springer, 2015.

\bibitem{sohl2015deep}
Jascha Sohl-Dickstein, Eric~A Weiss, Niru Maheswaranathan, and Surya Ganguli.
\newblock Deep unsupervised learning using nonequilibrium thermodynamics.
\newblock {\em arXiv preprint arXiv:1503.03585}, 2015.

\bibitem{song2020denoising}
Jiaming Song, Chenlin Meng, and Stefano Ermon.
\newblock Denoising diffusion implicit models.
\newblock {\em arXiv preprint arXiv:2010.02502}, 2020.

\bibitem{wolleb2022diffusion}
Julia Wolleb, Florentin Bieder, Robin Sandk{\"u}hler, and Philippe~C. Cattin.
\newblock Diffusion models for medical anomaly detection.
\newblock In {\em International Conference on Medical Image Computing and Computer-Assisted Intervention (MICCAI)}, pages 35--45. Springer, 2022.

\bibitem{wyatt2022anoddpm}
Julian Wyatt, Adam Leach, Sebastian~M. Schmon, and Chris~G. Willcocks.
\newblock Anoddpm: Anomaly detection with denoising diffusion probabilistic models using simplex noise.
\newblock In {\em Proceedings of the IEEE/CVF Conference on Computer Vision and Pattern Recognition (CVPR)}, pages 650--656. IEEE/CVF, 2022.

\bibitem{yin2015improved}
Chunyong Yin, Sun Zhang, Jin Wang, and Jeong-Uk Kim.
\newblock An improved k-means using in anomaly detection.
\newblock In {\em 2015 First International Conference on Computational Intelligence Theory, Systems and Applications (CITSA)}, pages 129--132. IEEE, 2015.

\bibitem{zhou2019beatgan}
Bin Zhou, Shenghua Liu, Bryan Hooi, Xueqi Cheng, and Jing Ye.
\newblock Beatgan: Anomalous rhythm detection using adversarially generated time series.
\newblock In {\em Proceedings of the Twenty-Eighth International Joint Conference on Artificial Intelligence (IJCAI)}, pages 4433--4439. IJCAI, 2019.

\bibitem{zhou2023conditional}
Haoxuan Zhou, Zihao Lei, Enrico Zio, Guangrui Wen, Zimin Liu, Yu~Su, and Xuefeng Chen.
\newblock Conditional feature disentanglement learning for anomaly detection in machines operating under time-varying conditions.
\newblock {\em Mechanical Systems and Signal Processing}, 191:110139, 2023.

\bibitem{zong2018deep}
Bo~Zong, Qi~Song, Martin~Renqiang Min, Wei Cheng, Cristian Lumezanu, Daeki Cho, and Haifeng Chen.
\newblock Deep autoencoding gaussian mixture model for unsupervised anomaly detection.
\newblock In {\em International Conference on Learning Representations (ICLR)}. ICLR, 2018.

\bibitem{zonta2020predictive}
Tiago Zonta, Cristiano~Andr{\'e} Da~Costa, Rodrigo da~Rosa~Righi, Miromar~Jose de~Lima, Eduardo~Silveira da~Trindade, and Guann~Pyng Li.
\newblock Predictive maintenance in the industry 4.0: A systematic literature review.
\newblock {\em Computers \& Industrial Engineering}, 150:106889, 2020.

\end{thebibliography}
	
\end{document}